%% file: main.tex
\pdfoutput=1

\documentclass{article} 
\usepackage{iclr2022_conference,times}

\input{math_commands.tex}

\usepackage{hyperref}
\usepackage{url}

\usepackage{todonotes}
\usepackage{xspace}
\usepackage{booktabs}
\newcommand\ie{i.\,e.\xspace}
\newcommand\eg{e.\,g.\xspace}

\title{Emergent Communication for Understanding Human Language Evolution: What's Missing?}

\author{Lukas Galke\\
Max Planck Institute for Psycholinguistics\\
Nijmegen, Netherlands\\
\texttt{lukas.galke@mpi.nl} \\
\And
Yoav Ram\\
School of Zoology\\
Faculty of Life Sciences \\
Sagol School of Neuroscience\\
Tel Aviv University, Israel \\
\texttt{yoav@yoavram.com} \\
\And
Limor Raviv\\
Max Planck Institute for Psycholinguistics\\
Nijmegen, Netherlands \\
Centre for Social, Cognitive and Affective Neuroscience\\
University of Glasgow, Scotland \\
\texttt{limor.raviv@mpi.nl} \\
}

\iclrfinalcopy{} 
\begin{document}

\maketitle

\begin{abstract}
    Emergent communication protocols among humans and artificial neural network agents do not yet share the same properties and show some critical mismatches in results. We describe three important phenomena with respect to the emergence and benefits of compositionality: ease-of-learning, generalization, and group size effects (i.e., larger groups create more systematic languages). The latter two are not fully replicated with neural agents, which hinders the use of neural emergent communication for language evolution research. We argue that one possible reason for these mismatches is that key cognitive and communicative constraints of humans are not yet integrated. Specifically, in humans, memory constraints and the alternation between the roles of speaker and listener underlie the emergence of linguistic structure, yet these constraints are typically absent in neural simulations. We suggest that introducing such communicative and cognitive constraints would promote more linguistically plausible behaviors with neural agents.
\end{abstract}

\section{Introduction}
Emergent communication has been widely studied in deep learning \citep{lazaridou2020emergent}, and in language evolution \citep{selten2007emergence,winters2015languages,raviv2019larger}. 
Both fields share communication games as a common experimental framework: a speaker describes an input, \eg, an object or a scene, and transmits a message to a listener, which has to guess or reconstruct the speaker's input. 
However, the languages of artificial neural network agents (neural agents) do not always exhibit the same linguistic properties as human languages. This presents a problem for using emergent communication as a model for language evolution of humans (or animals \citep{prat2019animals}). 

Here, we emphasize three important phenomena in human language evolution (described in detail in \Secref{sec:human-phenomena}) that relate to the emergence of compositional structure --- all of which have been discussed theoretically and confirmed experimentally with humans. First, compositional languages are easier to learn \citep{kirby2014iterated,carr2017cultural,raviv2021makes}.
Second, more compositionality allows for better generalization and facilitates convergence between strangers \citep{wray2007consequences,raviv2021makes}.
Third, larger populations generally develop more structured languages \citep{lupyan2010language,raviv2019larger}. 

However, in emergent communication between neural agents, two of the three phenomena are not yet replicated (see \Secref{sec:replication-attempts}). Although the ease-of-learning effect for compositional structure has been confirmed in multiple experiments \citep{DBLP:conf/nips/LiB19,guo2019emergence,DBLP:conf/acl/ChaabouniKBDB20}, recent work has shown that compositional structure is not necessary for generalization \citep{DBLP:conf/iclr/LazaridouHTC18,DBLP:conf/acl/ChaabouniKBDB20}.
Regarding the effect of group size, so far this could only be confirmed with continuous communication channels \citep{tieleman2019shaping}. With discrete communication, an increase in group size does not lead to the emergence of more compositional languages \citep{chaabouni2022emergent}. Only recently, \citet{rita2022on} have shown that the group size effect can be partially recovered by explicitly modeling population heterogeneity.

We propose two potential explanations for the striking mismatches between humans and neural agents. We argue that emergent communication simulations with neural networks typically lack two key communicative and cognitive factors that drive the emergence of compositionality in humans, and whose omission essentially eliminates the benefits of compositionality: memory constraints, and alternating roles. These are described in detail in \Secref{sec:reasons}, along with suggestions on how to tackle them, namely, limiting model capacity and sharing parameters within single agents.

\section{Important Properties of Human Languages}\label{sec:human-phenomena}
Two of the most fundamental properties of human languages are 1. a consistent form-to-meaning mapping and 2. a compositional structure \citep{hockett1960origin}.
Previous work suggests that compositional structure evolves as the trade-off between a compressability pressure (i.e., languages should be as simple and learnable as possible) and an expressivity pressure (i.e., languages should be able to successfully disambiguate between a variety of meanings) \citep{kirby2015compression}. 
Without a compressibility pressure, languages would become completely holistic (that is, with a unique symbol for each meaning) - which is highly expressive but poorly compressed. Without an expressivity pressure, languages would become degenerate (that is, comprised of a single symbol) - which is highly compressed but not expressive.
Yet, with both pressures present, as in the case of natural languages, structured languages with compositionality would emerge --- as these strike a balance between compressibility and expressivity. Notably, three phenomena related to compositionality have been demonstrated and discussed with humans:

\paragraph{Compositional languages are easier to learn.}
The driving force behind a compressibility pressure is that languages must be transmitted, learned, and used by multiple individuals, often from limited input and with limited exposure time \citep{smith2003complex}. 
Numerous iterated learning studies have shown that artificial languages become easier to learn over time \citep{kirby2008cumulative,winters2015languages,beckner2017emergence} --- a finding that is attributed to a simultaneous increase in compositionality. Indeed, artificial language learning experiments directly confirm that more compositional languages are learned better and faster by adults \citep{raviv2021makes}. 

\paragraph{Compositionality aids generalization.}
When a language exhibits a compositional structure, it is easier to generalize (describe new meanings) in a way that is transparent and understandable to other speakers \citep{kirby2002learning,zuidema2002poverty}. For example, the need to communicate over a growing number of different items in an open-ended meaning space promotes the emergence of more compositional systems \cite{carr2017cultural}. Recently, \citet{raviv2021makes} showed that compositional structure predicts generalization performance, with compositional languages that allow better generalizations that are also shared across different individuals .
The rationale is that humans cannot remember a holistic language (compressability pressure), and when they need to invent descriptions for new meanings, compositionality enables systematic and transparent composition of new label-meaning pairs from existing part-labels.

\paragraph{Larger groups create more compositional languages.}
Socio-demographic factors such as population size have long been assumed to be important determinants of language evolution \citep{wray2007consequences,nettle2012social,lupyan2010language}.
Specifically, global cross-linguistic studies found that bigger communities tend to have languages with more systematic and transparent structures \citep{lupyan2010language}. This result has also been experimentally confirmed, where larger groups of interacting participants created more compositional languages \citep{raviv2019larger}. 
These findings are attributed to compressibility pressures arising during communication: remembering partner-specific variants becomes increasingly more challenging as group size increases, and so memory constraints force larger groups to converge on more transparent languages.

\section{Replication Attempts with Neural Agents}\label{sec:replication-attempts}

Computational modeling has long been used to study language evolution \citep{kirby2002learning,smith2003complex,kirby2004ug,smith2008cultural,gong2008exploring,dale2012understanding,perfors2014language,kirby2015compression,steels2016agent}.
More recently, emergent communication with neural networks and reinforcement learning techniques was introduced~\citep{DBLP:conf/iclr/LazaridouPB17,DBLP:conf/nips/HavrylovT17,kottur-etal-2017-natural}.
Table~\ref{tab:experimental-settings} offers a (non-exhaustive) summary of recent emergent communication experiments with neural agents.
Both symbolic and visual inputs have been used \citep{DBLP:conf/iclr/LazaridouHTC18}, and the channel bandwidth (alphabet size to the power of message length) has been increasing with time. Most experiments use long short-term memory (LSTM)~\citep{lstm} as the agents' architecture, and training is most often carried out by the REINFORCE algorithm \citep{williams1992simple}. Crucially, most experiments have been limited to a pair of agents, although \citet{DBLP:conf/nips/LiB19} experimented with a single speaker and multiple listeners. Only recently, \citet{chaabouni2022emergent} and \citet{rita2022on} scaled up the group size. In all experiments except \cite{DBLP:conf/emnlp/GraesserCK19}, agents do not alternate between the roles of speaker and listener and instead employ distinct models for speaker and listener. For comparison, experiments with humans by \cite{raviv2019larger} have had $4 \times 16$ different visual meanings, a discrete communication channel with bandwidth $16^{16}$, and group sizes of 4-8 participants, who alternated between being speakers and listeners.

\begin{table}[ht]
    \caption{Neural emergent communication experiments. Input type and size of meaning space, channel bandwidth (alphabet size to the power of message length), task objective (reconstruction and discrimination, the latter with number of distractors), group size, role alternation between speaker and listener (RA), and presence of iterated learning (IL). Placeholders like N, K, or comma-separated lists mean that settings were varied.}
    \centering
    \scriptsize
    \begin{tabular}{lllllll}
    \toprule
    \textbf{Experiment}                         & \textbf{Inputs}           & \textbf{Channel}                            & \textbf{Objective} & \textbf{Groups} & \textbf{RA} & \textbf{IL} \\
         \midrule
          \citet{DBLP:conf/nips/HavrylovT17}    & visual (MSCOCO)           & $128^N$                                     & Discr. (128)       & $2$             & No          & No          \\
         \citet{kottur-etal-2017-natural}       & symbolic ($4^3$)          & $N^1$ & Reconstr.             & $2$             & No          & No          \\
         \citet{DBLP:conf/iclr/LazaridouPB17}   & pretrained visual         & $10^1$/$100^1$                              & Discr.             & $2$             & No          & No          \\
         \citet{DBLP:conf/iclr/LazaridouHTC18}  & symbolic ($463$)      & $10^2$/$17^5$/$40^{10}$                     & Discr. (5)         & $2$             & No          & No          \\
         \citet{DBLP:conf/iclr/LazaridouHTC18}  & visual ($124 \times 124$) & N/A                                         & Discr. (2,20)      & $2$             & No          & No          \\
          \citet{tieleman2019shaping}           & visual                    & \emph{continuous}                           & Reconstr.             & $1,2,4,8,16,32$             & No          & No          \\
          \citet{chaabouni2019anti}             & symbolic(K)               & $40^{30}$                                   & Reconstr.             & $50, 100$    & No          & No          \\
          \citet{DBLP:conf/emnlp/GraesserCK19}  & visual                    & $2^{8}$                                     & Discr.             & $N$             & Yes         & No          \\
          \citet{DBLP:conf/conll/RitaCD20}      & symbolic ($1000$)         & $40^{30}$                                   & Reconstr.           & $2$             & No          & No          \\
          \citet{DBLP:conf/nips/LiB19}          & symbolic ($8 \times 4$)   & $8^2$                                       & Discr. (5)         & $1:1,2,10$ & No          & Yes         \\
          \citet{kharitonov2020entropy}         & symbolic ($2^8$)          & $1$                                         & Partial Reconstr.     & $2$             & No          & No          \\
          \citet{kharitonov2020entropy}         & visual ($10^2$)           & $2^{10}$                                    & Discr. (100)        & $2$             & No          & No          \\
          \citet{DBLP:conf/acl/ChaabouniKBDB20} & symbolic ($2 \times 100$) & $100^3$                                     & Reconstr.             & $2$             & No          & Yes         \\
          \citet{chaabouni2022emergent}         & visual (ImageNet, CelebA) & $20^{10}$                                   & Discr. (20--1024)  & $2,20,100$             & No          & Yes         \\
          \citet{rita2022on}                    & symbolic ($4 \times 4$)   & $20^{10}$                                   & Reconstr.            & $2$,$10$             & No          & No          \\
      \bottomrule
    \end{tabular}
    \label{tab:experimental-settings}
\end{table}

With respect to the three linguistic phenomena described above for human participants, neural agents show a mixed pattern of results: 
First, ease-of-learning of compositional languages has been confirmed in emergent communication with neural agents \citep{guo2019emergence,DBLP:conf/acl/ChaabouniKBDB20,chaabouni2022emergent}. For example, more compositionality emerges when agents are being constantly reset and need to learn the language over and over again (similar to the iterated learning paradigm) \citep{DBLP:conf/nips/LiB19}. Similarly, \citet{guo2019emergence} found that compositional languages emerge in iterated learning experiments with neural agents because they are easier to learn. 

Second, in contrast to humans, neural agents can generalize well even without compositional communication protocols.
\cite{DBLP:conf/acl/ChaabouniKBDB20} have found that compositionality is \emph{not} necessary for generalization in neural agents (in line with earlier findings from \citet{DBLP:conf/iclr/LazaridouHTC18}). Although they argue that structure (however it emerges) prevails throughout evolution \emph{because} of its implied learnability advantage (in line with \citet{kirby2015compression}), the finding that compositionality aids generalization has nevertheless not been replicated with neural agents yet.

Third, in populations of autoencoders, where the encoder and decoders were decoupled and exchanged throughout training, larger communities produced representations with less idiosyncrasies \citep{tieleman2019shaping}. However, these experiments used a continuous, rather than discrete, channel, which has only recently been analyzed with an increasing population size \citep{chaabouni2022emergent,rita2022on}.
Although \citet{chaabouni2022emergent} argue that it is necessary to scale up emergent communication experiments to better align neural emergent communication with human language evolution, they have not found a systematic advantage of population size in generalization and ease-of-learning (in contrast with \citep{tieleman2019shaping}).
Similarly, \citet{rita2022on} found that language properties are not enhanced by population size alone. However, when adding heterogeneity through different learning rates, an increase in population size led to an increase in structure. 

\section{Potential Reasons for the Mismatch in Results}\label{sec:reasons}
Why does the population size not affect emergent neural communication? And why do neural agents not need compositionality to generalize? 
Our key argument is that crucial communicative and cognitive factors in humans have not yet been appropriately introduced to neural agents in emergent communication experiments. We argue that omitting these factors effectively removes the compressibility pressure that underlies the emergence and benefits of compositionality. In the following, we highlight two crucial factors: memory constraints and speaker/listener role alternation.

\paragraph{Memory Constraints}

Human memory limitation is one of the most important constraints of language learning, and underlies the compressibility pressure in language use and transmission. Indeed, sequence memory constraints induce structure emergence in iterated learning \cite{cornish2017sequence}, and underlie group size effects in real-world settings \citep{meir2012influence,wray2007consequences} and in communication games with humans  \citep{raviv2019larger}, where more compositionality is promoted because individuals cannot memorize partner-specific variations as the group size increases. In contrast, a key ingredient for the success of neural networks is their overparameterization \citep{nakkiran2021deep}, which means that their capacity is in fact sufficient to completely "memorize" sender-specific idiosyncratic languages. We propose that this overparameterization significantly reduces compressibility pressure, effectively eliminating the potential benefits of compositional structure for learning and generalization. 

Therefore, we suggest that in communication games, the model capacity, \ie, how much information the model can store (well quantifiable, \eg, see \cite{mackay2003information}), should be considered in relation to the number of different meanings and the space of all possible messages, \ie alphabet size to the power of (maximum) message length $|\gA|^L$.
We hypothesize that for compositionality to emerge, the model capacity 
needs to be less than required to memorize a separate message for each meaning from every agent, but also less than the theoretical channel bandwidth, such that it becomes necessary to reuse substructures within the messages. 
Consistent with our position, \citet{DBLP:conf/atal/Resnick0FDC20} and \citet{gupta-etal-2020-compositionality} verified that learning an underlying compositional structure requires less capacity than memorizing a dataset. Similarly to us, the authors of both works assume a range in model capacity that facilitates compositionality, but so far only determine a lower bound, while we argue here about the upper bound(s). 

\paragraph{Role Alternation}
In current neural emergent communication experiments, one agent generates the message, and the other only processes it. This in sharp contrast to human communication, where speakers can reproduce any linguistic message that they can understand \citep{hockett1960origin}. Indeed, dyadic and group communication experiments with humans typically have people alternating between the roles of speaker and listener throughout the experiment \citep{kirby2015compression,raviv2019larger,motamedi2021emergence}. This is only rarely reflected in work with neural agents (Table~\ref{tab:experimental-settings}).

A straight-forward implementation of role alternation is to have shared parameters within the (generative) speaker role and the (discriminative) listener role of the same neural agent. This would make the architecture of neural agents more similar to the human brain, where shared neural mechanisms support both the production and the comprehension of natural speech \citep{SilbertE4687}. One way to do this would be to tie the output layer's weights to input embedding, a well known concept in language modeling \citep{word2vec,t5}. Some experiments already implement role alternation, \eg, in multi-agent communication with given language descriptions \citep{DBLP:conf/emnlp/GraesserCK19},
or in language acquisition from image captions where agents simultaneously learn by cognition and production \citep{nikolaus-fourtassi-2021-modeling}. We suggest role alternation should also implemented in emergent communication experiments to ensure more linguistically plausible dynamics.

\section{Conclusion}
We have outlined important discrepancies in the results between emergent communication with human versus neural agents and suggested that these can be explained by the absence of key cognitive and communicative factors that drive human language evolution: memory constraints and speaker-listener role alternation.
We suggest that including these factors in future work would mimic the compressability pressure and compositionality benefits observed with human agents, and consequentially would make emergent neural communication protocols more linguistically plausible. 
Notably, additional psycho- and socio-linguistic factors may affect language evolution, and might also play a role in explaining the discrepancy in results. 

\section*{Acknowledgements}
We thank Mitja Nikolaus for valuable discussions on role alternation and parameter sharing.
This research is partly funded by Minerva Center for Lab Evolution; John Templeton Foundation grant.

\bibliography{custom}
\bibliographystyle{iclr2022_conference}

\end{document}

%% file: math_commands.tex
\usepackage{amsmath,amsfonts,bm}

\def\Secref#1{Section~\ref{#1}}
\def\eqref#1{equation~\ref{#1}}
\def\1{\bm{1}}

\DeclareMathAlphabet{\mathsfit}{\encodingdefault}{\sfdefault}{m}{sl}
\SetMathAlphabet{\mathsfit}{bold}{\encodingdefault}{\sfdefault}{bx}{n}

\def\gA{{\mathcal{A}}}

%% file: main.bbl
\begin{thebibliography}{51}
\providecommand{\natexlab}[1]{#1}
\providecommand{\url}[1]{\texttt{#1}}
\expandafter\ifx\csname urlstyle\endcsname\relax
  \providecommand{\doi}[1]{doi: #1}\else
  \providecommand{\doi}{doi: \begingroup \urlstyle{rm}\Url}\fi

\bibitem[Beckner et~al.(2017)Beckner, Pierrehumbert, and
  Hay]{beckner2017emergence}
Clay Beckner, Janet~B Pierrehumbert, and Jennifer Hay.
\newblock The emergence of linguistic structure in an online iterated learning
  task.
\newblock \emph{Journal of Language Evolution}, 2\penalty0 (2):\penalty0
  160--176, 2017.

\bibitem[Carr et~al.(2017)Carr, Smith, Cornish, and Kirby]{carr2017cultural}
Jon~W Carr, Kenny Smith, Hannah Cornish, and Simon Kirby.
\newblock The cultural evolution of structured languages in an open-ended,
  continuous world.
\newblock \emph{Cognitive science}, 41\penalty0 (4):\penalty0 892--923, 2017.

\bibitem[Chaabouni et~al.(2019)Chaabouni, Kharitonov, Dupoux, and
  Baroni]{chaabouni2019anti}
Rahma Chaabouni, Eugene Kharitonov, Emmanuel Dupoux, and Marco Baroni.
\newblock Anti-efficient encoding in emergent communication.
\newblock In \emph{NeurIPS}, pp.\  6290--6300, 2019.

\bibitem[Chaabouni et~al.(2020)Chaabouni, Kharitonov, Bouchacourt, Dupoux, and
  Baroni]{DBLP:conf/acl/ChaabouniKBDB20}
Rahma Chaabouni, Eugene Kharitonov, Diane Bouchacourt, Emmanuel Dupoux, and
  Marco Baroni.
\newblock Compositionality and generalization in emergent languages.
\newblock In \emph{{ACL}}, pp.\  4427--4442. Association for Computational
  Linguistics, 2020.

\bibitem[Chaabouni et~al.(2022)Chaabouni, Strub, Altch{\'e}, Tarassov, Tallec,
  Davoodi, Mathewson, Tieleman, Lazaridou, and Piot]{chaabouni2022emergent}
Rahma Chaabouni, Florian Strub, Florent Altch{\'e}, Eugene Tarassov, Corentin
  Tallec, Elnaz Davoodi, Kory~Wallace Mathewson, Olivier Tieleman, Angeliki
  Lazaridou, and Bilal Piot.
\newblock Emergent communication at scale.
\newblock In \emph{ICLR}, 2022.
\newblock URL \url{https://openreview.net/forum?id=AUGBfDIV9rL}.

\bibitem[Cornish et~al.(2017)Cornish, Dale, Kirby, and
  Christiansen]{cornish2017sequence}
Hannah Cornish, Rick Dale, Simon Kirby, and Morten~H Christiansen.
\newblock Sequence memory constraints give rise to language-like structure
  through iterated learning.
\newblock \emph{PloS one}, 12\penalty0 (1):\penalty0 e0168532, 2017.

\bibitem[Dale \& Lupyan(2012)Dale and Lupyan]{dale2012understanding}
Rick Dale and Gary Lupyan.
\newblock Understanding the origins of morphological diversity: The linguistic
  niche hypothesis.
\newblock \emph{Advances in complex systems}, 15\penalty0 (03n04):\penalty0
  1150017, 2012.

\bibitem[Gong et~al.(2008)Gong, Minett, and Wang]{gong2008exploring}
Tao Gong, James~W Minett, and William S-Y Wang.
\newblock Exploring social structure effect on language evolution based on a
  computational model.
\newblock \emph{Connection Science}, 20\penalty0 (2-3):\penalty0 135--153,
  2008.

\bibitem[Graesser et~al.(2019)Graesser, Cho, and
  Kiela]{DBLP:conf/emnlp/GraesserCK19}
Laura Graesser, Kyunghyun Cho, and Douwe Kiela.
\newblock Emergent linguistic phenomena in multi-agent communication games.
\newblock In \emph{{EMNLP/IJCNLP} {(1)}}, pp.\  3698--3708. Association for
  Computational Linguistics, 2019.

\bibitem[Guo et~al.(2019)Guo, Ren, Havrylov, Frank, Titov, and
  Smith]{guo2019emergence}
Shangmin Guo, Yi~Ren, Serhii Havrylov, Stella Frank, Ivan Titov, and Kenny
  Smith.
\newblock The emergence of compositional languages for numeric concepts through
  iterated learning in neural agents.
\newblock \emph{CoRR}, abs/1910.05291, 2019.

\bibitem[Gupta et~al.(2020)Gupta, Resnick, Foerster, Dai, and
  Cho]{gupta-etal-2020-compositionality}
Abhinav Gupta, Cinjon Resnick, Jakob Foerster, Andrew Dai, and Kyunghyun Cho.
\newblock Compositionality and capacity in emergent languages.
\newblock In \emph{Proceedings of the 5th Workshop on Representation Learning
  for NLP}, pp.\  34--38, Online, July 2020. Association for Computational
  Linguistics.
\newblock \doi{10.18653/v1/2020.repl4nlp-1.5}.
\newblock URL \url{https://aclanthology.org/2020.repl4nlp-1.5}.

\bibitem[Havrylov \& Titov(2017)Havrylov and Titov]{DBLP:conf/nips/HavrylovT17}
Serhii Havrylov and Ivan Titov.
\newblock Emergence of language with multi-agent games: Learning to communicate
  with sequences of symbols.
\newblock In \emph{{NeurIPS}}, pp.\  2149--2159, 2017.

\bibitem[Hochreiter \& Schmidhuber(1997)Hochreiter and Schmidhuber]{lstm}
Sepp Hochreiter and Jürgen Schmidhuber.
\newblock Long short-term memory.
\newblock \emph{Neural Computation}, 9\penalty0 (8):\penalty0 1735--1780, 1997.
\newblock \doi{10.1162/neco.1997.9.8.1735}.

\bibitem[Hockett(1960)]{hockett1960origin}
Charles~F Hockett.
\newblock The origin of speech.
\newblock \emph{Scientific American}, 203\penalty0 (3):\penalty0 88--97, 1960.

\bibitem[Kharitonov et~al.(2020)Kharitonov, Chaabouni, Bouchacourt, and
  Baroni]{kharitonov2020entropy}
Eugene Kharitonov, Rahma Chaabouni, Diane Bouchacourt, and Marco Baroni.
\newblock Entropy minimization in emergent languages, 2020.

\bibitem[Kirby(2002)]{kirby2002learning}
Simon Kirby.
\newblock Learning, bottlenecks and the evolution of recursive syntax.
\newblock 2002.

\bibitem[Kirby et~al.(2004)Kirby, Smith, and Brighton]{kirby2004ug}
Simon Kirby, Kenny Smith, and Henry Brighton.
\newblock From ug to universals: Linguistic adaptation through iterated
  learning.
\newblock \emph{Studies in Language. International Journal sponsored by the
  Foundation “Foundations of Language”}, 28\penalty0 (3):\penalty0
  587--607, 2004.

\bibitem[Kirby et~al.(2008)Kirby, Cornish, and Smith]{kirby2008cumulative}
Simon Kirby, Hannah Cornish, and Kenny Smith.
\newblock Cumulative cultural evolution in the laboratory: An experimental
  approach to the origins of structure in human language.
\newblock \emph{Proceedings of the National Academy of Sciences}, 105\penalty0
  (31):\penalty0 10681--10686, 2008.

\bibitem[Kirby et~al.(2014)Kirby, Griffiths, and Smith]{kirby2014iterated}
Simon Kirby, Tom Griffiths, and Kenny Smith.
\newblock Iterated learning and the evolution of language.
\newblock \emph{Current opinion in neurobiology}, 28:\penalty0 108--114, 2014.

\bibitem[Kirby et~al.(2015)Kirby, Tamariz, Cornish, and
  Smith]{kirby2015compression}
Simon Kirby, Monica Tamariz, Hannah Cornish, and Kenny Smith.
\newblock Compression and communication in the cultural evolution of linguistic
  structure.
\newblock \emph{Cognition}, 141:\penalty0 87--102, 2015.

\bibitem[Kottur et~al.(2017)Kottur, Moura, Lee, and
  Batra]{kottur-etal-2017-natural}
Satwik Kottur, Jos{\'e} Moura, Stefan Lee, and Dhruv Batra.
\newblock Natural language does not emerge {`}naturally{'} in multi-agent
  dialog.
\newblock In \emph{Proceedings of the 2017 Conference on Empirical Methods in
  Natural Language Processing}, pp.\  2962--2967, Copenhagen, Denmark,
  September 2017. Association for Computational Linguistics.
\newblock \doi{10.18653/v1/D17-1321}.
\newblock URL \url{https://aclanthology.org/D17-1321}.

\bibitem[Lazaridou \& Baroni(2020)Lazaridou and Baroni]{lazaridou2020emergent}
Angeliki Lazaridou and Marco Baroni.
\newblock Emergent multi-agent communication in the deep learning era.
\newblock \emph{CoRR}, abs/2006.02419, 2020.

\bibitem[Lazaridou et~al.(2017)Lazaridou, Peysakhovich, and
  Baroni]{DBLP:conf/iclr/LazaridouPB17}
Angeliki Lazaridou, Alexander Peysakhovich, and Marco Baroni.
\newblock Multi-agent cooperation and the emergence of (natural) language.
\newblock In \emph{{ICLR}}. OpenReview.net, 2017.

\bibitem[Lazaridou et~al.(2018)Lazaridou, Hermann, Tuyls, and
  Clark]{DBLP:conf/iclr/LazaridouHTC18}
Angeliki Lazaridou, Karl~Moritz Hermann, Karl Tuyls, and Stephen Clark.
\newblock Emergence of linguistic communication from referential games with
  symbolic and pixel input.
\newblock In \emph{{ICLR}}. OpenReview.net, 2018.

\bibitem[Li \& Bowling(2019)Li and Bowling]{DBLP:conf/nips/LiB19}
Fushan Li and Michael Bowling.
\newblock Ease-of-teaching and language structure from emergent communication.
\newblock In \emph{NeurIPS}, pp.\  15825--15835, 2019.

\bibitem[Lupyan \& Dale(2010)Lupyan and Dale]{lupyan2010language}
Gary Lupyan and Rick Dale.
\newblock Language structure is partly determined by social structure.
\newblock \emph{PloS one}, 5\penalty0 (1):\penalty0 e8559, 2010.

\bibitem[MacKay et~al.(2003)MacKay, Mac~Kay, et~al.]{mackay2003information}
David~JC MacKay, David~JC Mac~Kay, et~al.
\newblock \emph{Information theory, inference and learning algorithms}.
\newblock Cambridge university press, 2003.

\bibitem[Meir et~al.(2012)Meir, Israel, Sandler, Padden, and
  Aronoff]{meir2012influence}
Irit Meir, Assaf Israel, Wendy Sandler, Carol~A Padden, and Mark Aronoff.
\newblock The influence of community on language structure: evidence from two
  young sign languages.
\newblock \emph{Linguistic Variation}, 12\penalty0 (2):\penalty0 247--291,
  2012.

\bibitem[Mikolov et~al.(2013)Mikolov, Sutskever, Chen, Corrado, and
  Dean]{word2vec}
Tom{\'{a}}s Mikolov, Ilya Sutskever, Kai Chen, Gregory~S. Corrado, and Jeffrey
  Dean.
\newblock Distributed representations of words and phrases and their
  compositionality.
\newblock In \emph{{NIPS}}, pp.\  3111--3119, 2013.

\bibitem[Motamedi et~al.(2021)Motamedi, Smith, Schouwstra, Culbertson, and
  Kirby]{motamedi2021emergence}
Yasamin Motamedi, Kenny Smith, Marieke Schouwstra, Jennifer Culbertson, and
  Simon Kirby.
\newblock The emergence of systematic argument distinctions in artificial sign
  languages.
\newblock \emph{Journal of Language Evolution}, 6\penalty0 (2):\penalty0
  77--98, 2021.

\bibitem[Nakkiran et~al.(2021)Nakkiran, Kaplun, Bansal, Yang, Barak, and
  Sutskever]{nakkiran2021deep}
Preetum Nakkiran, Gal Kaplun, Yamini Bansal, Tristan Yang, Boaz Barak, and Ilya
  Sutskever.
\newblock Deep double descent: Where bigger models and more data hurt.
\newblock \emph{Journal of Statistical Mechanics: Theory and Experiment},
  2021\penalty0 (12):\penalty0 124003, 2021.

\bibitem[Nettle(2012)]{nettle2012social}
Daniel Nettle.
\newblock Social scale and structural complexity in human languages.
\newblock \emph{Philosophical Transactions of the Royal Society B: Biological
  Sciences}, 367\penalty0 (1597):\penalty0 1829--1836, 2012.

\bibitem[Nikolaus \& Fourtassi(2021)Nikolaus and
  Fourtassi]{nikolaus-fourtassi-2021-modeling}
Mitja Nikolaus and Abdellah Fourtassi.
\newblock Modeling the interaction between perception-based and
  production-based learning in children{'}s early acquisition of semantic
  knowledge.
\newblock In \emph{Proceedings of the 25th Conference on Computational Natural
  Language Learning}, pp.\  391--407, Online, November 2021. Association for
  Computational Linguistics.
\newblock \doi{10.18653/v1/2021.conll-1.31}.
\newblock URL \url{https://aclanthology.org/2021.conll-1.31}.

\bibitem[Perfors \& Navarro(2014)Perfors and Navarro]{perfors2014language}
Andrew Perfors and Daniel~J Navarro.
\newblock Language evolution can be shaped by the structure of the world.
\newblock \emph{Cognitive science}, 38\penalty0 (4):\penalty0 775--793, 2014.

\bibitem[Prat(2019)]{prat2019animals}
Yosef Prat.
\newblock Animals have no language, and humans are animals too.
\newblock \emph{Perspectives on Psychological Science}, 14\penalty0
  (5):\penalty0 885--893, 2019.

\bibitem[Raffel et~al.(2020)Raffel, Shazeer, Roberts, Lee, Narang, Matena,
  Zhou, Li, and Liu]{t5}
Colin Raffel, Noam Shazeer, Adam Roberts, Katherine Lee, Sharan Narang, Michael
  Matena, Yanqi Zhou, Wei Li, and Peter~J. Liu.
\newblock Exploring the limits of transfer learning with a unified text-to-text
  transformer.
\newblock \emph{J. Mach. Learn. Res.}, 21:\penalty0 140:1--140:67, 2020.

\bibitem[Raviv et~al.(2019)Raviv, Meyer, and Lev-Ari]{raviv2019larger}
Limor Raviv, Antje Meyer, and Shiri Lev-Ari.
\newblock Larger communities create more systematic languages.
\newblock \emph{Proceedings of the Royal Society B}, 286\penalty0
  (1907):\penalty0 20191262, 2019.

\bibitem[Raviv et~al.(2021)Raviv, de~Heer~Kloots, and Meyer]{raviv2021makes}
Limor Raviv, Marianne de~Heer~Kloots, and Antje Meyer.
\newblock What makes a language easy to learn? a preregistered study on how
  systematic structure and community size affect language learnability.
\newblock \emph{Cognition}, 210:\penalty0 104620, 2021.

\bibitem[Resnick et~al.(2020)Resnick, Gupta, Foerster, Dai, and
  Cho]{DBLP:conf/atal/Resnick0FDC20}
Cinjon Resnick, Abhinav Gupta, Jakob~N. Foerster, Andrew~M. Dai, and Kyunghyun
  Cho.
\newblock Capacity, bandwidth, and compositionality in emergent language
  learning.
\newblock In \emph{{AAMAS}}, pp.\  1125--1133. International Foundation for
  Autonomous Agents and Multiagent Systems, 2020.

\bibitem[Rita et~al.(2020)Rita, Chaabouni, and
  Dupoux]{DBLP:conf/conll/RitaCD20}
Mathieu Rita, Rahma Chaabouni, and Emmanuel Dupoux.
\newblock "lazimpa": Lazy and impatient neural agents learn to communicate
  efficiently.
\newblock In \emph{CoNLL}, pp.\  335--343. Association for Computational
  Linguistics, 2020.

\bibitem[Rita et~al.(2022)Rita, Strub, Grill, Pietquin, and Dupoux]{rita2022on}
Mathieu Rita, Florian Strub, Jean-Bastien Grill, Olivier Pietquin, and Emmanuel
  Dupoux.
\newblock On the role of population heterogeneity in emergent communication.
\newblock In \emph{ICLR}, 2022.
\newblock URL \url{https://openreview.net/forum?id=5Qkd7-bZfI}.

\bibitem[Selten \& Warglien(2007)Selten and Warglien]{selten2007emergence}
Reinhard Selten and Massimo Warglien.
\newblock The emergence of simple languages in an experimental coordination
  game.
\newblock \emph{Proceedings of the National Academy of Sciences}, 104\penalty0
  (18):\penalty0 7361--7366, 2007.

\bibitem[Silbert et~al.(2014)Silbert, Honey, Simony, Poeppel, and
  Hasson]{SilbertE4687}
Lauren~J. Silbert, Christopher~J. Honey, Erez Simony, David Poeppel, and Uri
  Hasson.
\newblock Coupled neural systems underlie the production and comprehension of
  naturalistic narrative speech.
\newblock \emph{Proceedings of the National Academy of Sciences}, 111\penalty0
  (43):\penalty0 E4687--E4696, 2014.
\newblock ISSN 0027-8424.
\newblock \doi{10.1073/pnas.1323812111}.
\newblock URL \url{https://www.pnas.org/content/111/43/E4687}.

\bibitem[Smith \& Kirby(2008)Smith and Kirby]{smith2008cultural}
Kenny Smith and Simon Kirby.
\newblock Cultural evolution: implications for understanding the human language
  faculty and its evolution.
\newblock \emph{Philosophical Transactions of the Royal Society B: Biological
  Sciences}, 363\penalty0 (1509):\penalty0 3591--3603, 2008.

\bibitem[Smith et~al.(2003)Smith, Brighton, and Kirby]{smith2003complex}
Kenny Smith, Henry Brighton, and Simon Kirby.
\newblock Complex systems in language evolution: the cultural emergence of
  compositional structure.
\newblock \emph{Advances in complex systems}, 6\penalty0 (04):\penalty0
  537--558, 2003.

\bibitem[Steels(2016)]{steels2016agent}
Luc Steels.
\newblock Agent-based models for the emergence and evolution of grammar.
\newblock \emph{Philosophical Transactions of the Royal Society B: Biological
  Sciences}, 371\penalty0 (1701):\penalty0 20150447, 2016.

\bibitem[Tieleman et~al.(2019)Tieleman, Lazaridou, Mourad, Blundell, and
  Precup]{tieleman2019shaping}
Olivier Tieleman, Angeliki Lazaridou, Shibl Mourad, Charles Blundell, and Doina
  Precup.
\newblock Shaping representations through communication: community size effect
  in artificial learning systems.
\newblock \emph{CoRR}, abs/1912.06208, 2019.

\bibitem[Williams(1992)]{williams1992simple}
Ronald~J Williams.
\newblock Simple statistical gradient-following algorithms for connectionist
  reinforcement learning.
\newblock \emph{Machine learning}, 8\penalty0 (3):\penalty0 229--256, 1992.

\bibitem[Winters et~al.(2015)Winters, Kirby, and Smith]{winters2015languages}
James Winters, Simon Kirby, and Kenny Smith.
\newblock Languages adapt to their contextual niche.
\newblock \emph{Language and Cognition}, 7\penalty0 (3):\penalty0 415--449,
  2015.

\bibitem[Wray \& Grace(2007)Wray and Grace]{wray2007consequences}
Alison Wray and George~W Grace.
\newblock The consequences of talking to strangers: Evolutionary corollaries of
  socio-cultural influences on linguistic form.
\newblock \emph{Lingua}, 117\penalty0 (3):\penalty0 543--578, 2007.

\bibitem[Zuidema(2002)]{zuidema2002poverty}
Willem Zuidema.
\newblock How the poverty of the stimulus solves the poverty of the stimulus.
\newblock \emph{Advances in neural information processing systems}, 15, 2002.

\end{thebibliography}
